Title Page

**Title: Heart Rate and its Variability from Short-term ECG Recordings as Biomarkers for Detecting Mild Cognitive Impairment in Indian Population**

**Running title:** Heart Rate and Heart Rate Variability as biomarker for MCI

**Author list:**

1. Anjo Xavier, M.Tech. [1]
2. Sneha Noble, M.Tech. [1]
3. Justin Joseph, Ph.D. [1]
4. Thomas Gregor Issac*, MD-PhD, Associate Professor [1]

**Author affiliation**

1. Centre for Brain Research, Indian institute of Science, Bangalore

**Author of Correspondence:** Prof. Thomas Gregor Issac

ORCID ID: 0000-0003-3148-3466

Telephone: 080-22933741

Email: thomasgregor@cbr-iisc.ac.in

Mailing address: Centre for Brain Research,

Indian Institute of Science Campus, CV Raman Avenue,

Bangalore-560012. India.

# Heart Rate and its Variability from Short-term ECG Recordings as Biomarkers for Detecting Mild Cognitive Impairment in Indian Population

**Abstract**

Alterations in Heart Rate (HR) and Heart Rate Variability (HRV) can reflect autonomic dysfunction associated with neurodegeneration. We investigate the influence of Mild Cognitive Impairment (MCI) on HR and its variability measures in the Indian population by designing a complete signal processing pipeline to detect the R-wave peaks and compute HR and HRV features from ECG recordings of 10 seconds, for point-of-care applications. The study cohort involves 297 urban participants, among which 48.48% are male and 51.51% are female. From the Addenbrooke's Cognitive Examination-III (ACE-III), MCI is detected in 19.19% of participants and the rest, 80.8% of them are cognitively healthy. Statistical features like central tendency (mean and root mean square (RMS) of the Normal-to-Normal (NN) intervals) and dispersion (standard deviation (SD) of all NN intervals (SDNN) and root mean square of successive differences of NN intervals (RMSSD)) of beat-to-beat intervals are computed. The Wilcoxon rank sum test reveals that mean of NN intervals ($p = 0.0021$), the RMS of NN intervals ($p = 0.0014$), the SDNN ($p = 0.0192$) and the RMSSD ($p = 0.0206$) values differ significantly between MCI and non-MCI classes, for a level of significance, 0.05. Machine learning classifiers like, Support Vector Machine (SVM), Discriminant Analysis (DA) and Naive Bayes (NB) driven by mean NN intervals, RMS, SDNN and RMSSD, show a high accuracy of 80.80% on each individual feature input. Individuals with MCI are observed to have comparatively higher HR than healthy subjects. HR and its variability can be considered as potential biomarkers for detecting MCI.

# 1. Introduction

Mild Cognitive Impairment (MCI) is a prodromal stage of dementia, characterized by cognitive decline beyond that expected for age but without significantly impacting daily activities (Taragano et al., 2008). It is considered as an intermediate state between normal cognitive aging and dementia. Early diagnosis of MCI is crucial for planning potential interventions and management strategies (Eshkoor et al., 2015). The prevalence of MCI increases with age in both genders. Patel et al., (2018) has reported that there is a prevalence of 15% in elderly people of age greater than 80 years, to 2.6% in people of age less than 60 years in India.

There is sufficient evidence for the association between the Heart Rate Variability (HRV) and MCI (Boudaya et al., 2024, Dalise et al., 2020). The HRV refers to the variation in time elapsed between successive heartbeats. It is a feasible, non-invasive measure of influence of the Autonomic Nervous system (ANS) on heart rate regulation and it reflects the dynamic balance between sympathetic and parasympathetic nervous system activity (Collins et al., 2012). Derby et al. (2023), has proved that HRV is associated with global cognition and cognitive impairment. Arakaki et al. (2023) showed the evidence that changes in Heart Rate (HR) and HRV may serve as indicators of cognitive decline and the progression of neurodegenerative disorders. Longitudinal studies (Nicolini et al., 2024) have found associations between baseline HRV measures and subsequent cognitive decline or dementia diagnosis. Altered or reduced HRV has been associated with different factors like autonomic dysfunction (Bassi and Marco, 2015), cardiovascular risk factors (Kulshreshtha et al., 2017, Livingston et al., 2020), inflammation, and oxidative stress, which ultimately affects the brain-heart axis.

Liu et al. (2023) have examined that AD patients with agitation (presence) have a lower HRV and/or a steeper fall in HRV over time than those without agitation. Imahori et al. (2022) have illustrated that there is an increased risk of dementia when the RHR is greater than or equal to 80 beats per minute, regardless of vascular risk factors and underlying cardio-vascular diseases (CVDs). Similarly, the participants with RHR greater than or equal to 70 bpm experienced a faster deterioration in general cognitive function than those with RHR of 60–69 bpm.

Deng et al. (2022) conducted a comprehensive logistic regression analysis, revealing a significant association between RHR and cognitive function. Elevated RHR values, predominantly exceeding 80 beats per minute, were frequently observed among individuals experiencing cognitive decline. The study included 4177 participants, encompassing 2354 cases of Alzheimer's Disease (AD) and 989 instances of vascular dementia (VaD). Similarly, Kim et al. (2022) found an association between increased RHR and all combined types of dementia in a meta-analysis. They suggest that, in order to ascertain whether elevated RHR is a direct cause of cognitive deterioration, large-scale prospective cohort studies are necessary. Molloy et al. (2023) found an association between higher RHR and lower Mini-Mental State Examination (MMSE) scores among individuals with pathological amyloid/tau ratios in the cerebrospinal fluid.

Zeki et al. (2014) also reported a cross-sectional association between HRV and cognitive performance via a study conducted among community-dwelling elderly Mexican Americans, an ethnic group at high risk for cardiovascular disease risk factors. Reduced HRV was reported to be associated with worse performance on the global test of cognitive function. Grässler et al., (2023) showed that MCI patients have significantly lower HRV compared to the healthy controls (HC) at resting state.

Although there have been certain studies that unwrap the correlation between HR, HRV and cognitive performance at global scale (Tsunoda et al., 2017), the applicability of the association on Indian elderly population with MCI is unknown. The objective of this study is to investigate the correlation between HR, HRV, and cognitive decline, with a specific emphasis on those diagnosed with MCI within the elderly Indian population. Our aim is to provide a comprehensive analysis of the potential implications of HR and HRV as biomarkers for early identification of cognitive impairment. We develop a signal processing pipeline that can detect R-wave peaks and calculate HR and HRV from brief ECG recordings, for point-of-care/bedside applications.

## 2. Methodology

### 2.1 Participants

The present study was conducted as part of the Centre for Brain Research-Tata Longitudinal Study for Ageing (CBR-TLSA) and approved by the Institutional Ethics Committee. The study protocol follows all necessary ethical guidelines including participants' safety, confidentiality, informed consent and voluntary participation. It focuses on participants aged 45 years and older, residing in urban Bangalore, India. The cohort underwent comprehensive clinical, biochemical, and radiological assessments at regular intervals. The recruitment process for the cohort involved stringent exclusion criteria to ensure the robustness of the study. Participants with terminal medical illnesses, psychiatric disorders, severe visual or hearing impairments, or locomotor disabilities likely to interfere with study assessments were excluded from the cohort. The selected participants were typically from working, middle-class backgrounds, demonstrating a reasonable level of literacy and having undergone significant lifestyle changes over the past few decades.

The study participants were recruited from the CBR-TLSA cohort. A total of 600 participants were initially enrolled in this investigation. However, data from 536 participants were only utilised for the current study, as the remaining participants' data lacked adequate ECG waveforms or had missing cognitive scores. Among the cohort of 536 participants, 239 individuals were subsequently excluded from analysis based on their concurrent use of antihypertensive medications, given the potential influence of such medications on both HR and HRV measures which led to a final cohort size of 297. The mean age of the participants in the cohort was 61.42 ± 9.75 years. Out of the 297 individuals, 48.48% (144) were male, and 51.51% (153) were female.

**2.2 Cognitive Assessment**

The Addenbrooke's Cognitive Examination (ACE-III) is used to assess cognitive function of the participants and identify the MCI cases. Generally, the ACE is used to aid in the detection and evaluation of cognitive impairment, including dementia and MCI. The ACE test is commonly used in clinical practice and research settings to assess cognitive function quickly and reliably. The ACE-III is a revised version of the original ACE test, incorporating additional subtests and enhancements to improve its sensitivity and specificity in detecting cognitive impairment. The ACE-III provides a total score as well as scores for individual cognitive domains, allowing for a comprehensive assessment of cognitive function (Takenoshita et al., 2019).

As already mentioned, in the current study, the cognitive assessment and group classification are performed using the ACE-III test, consisting of 24 diverse questions encompassing various cognitive domains, such as attention, memory, fluency, language, and visuospatial abilities. Participants scoring below 88 for the ACE-III test were classified as MCI, whereas those scoring greater than or equal to 88 were classified as healthy controls (Takenoshita et al., 2019). The comprehensive use of the ACE-III test facilitated the identification of individuals experiencing cognitive decline, contributing valuable insights towards the association between HR dynamics, HRV, and cognitive impairments. Among the subjects, 19.19% of the participants (57 in number) showed MCI, leaving the remaining 80.8% (240) classified as cognitively normal. The age spectrum ranged from 45 to 92 years, with a calculated mean age of 61.42 ± 9.75 years. Specifically, the mean age among individuals with MCI was 61.75 ± 10.19 years, and that for non-MCI individuals was 61.34 ± 9.66 years.

**2.3 Computation of HR and HRV Features**

Classical 12-lead ECG recordings are collected from the participants. The ECG recordings are obtained using the BPL Cardiart 9108 12-channel ECG recorder, operating at a high sampling frequency of 16 kHz. During the recording session, the participants were in a supine position. ECG data were collected for a standardized duration of 10 seconds. Compared to other studies, this method utilizes a short duration of ECG acquisition. From the acquired signal, the Lead II was used further as the R wave peak is more prominent and isolated in the lead II. Subsequently, preprocessing is done. In the preprocessing stage, the signal undergoes detrending. The detrending eliminates low-frequency trends or baseline drifts. After detrending, a band-pass filter (0.5 to 100 Hz) is applied to extract the spectral components of diagnostic importance and avoid radio frequency (RF) interferences. A notch filter is used in addition to the band pass filter to avoid powerline interference.

After the preprocessing stage, R-wave peaks are identified using a customized peak-detection algorithm. A time series of NN intervals is generated from the time elapsed between successive R-wave peaks. Further, statistical features representing central tendency and dispersion are extracted from the NN intervals. The measures of central tendency include mean and root mean square (RMS), while measures of dispersion encompass standard Deviation (SD) of all NN intervals (SDNN) and RMS of successive differences of NN intervals (RMSSD).

The mean NN interval is.

$$\mu_{NN} = \frac{1}{N}\sum_{i=1}^{N} NN_i \qquad (1)$$

In (1), $NN_i$ is an arbitrary NN interval and $N$ is the total number of NN interval values in the time series. The RMS value of the NN intervals is,

$$\rho_{NN} = \sqrt{\frac{1}{N}\sum_{i=1}^{N}(NN_i)^2} \qquad (2)$$

The standard deviation of the NN intervals is,

$$\sigma_{NN} = \sqrt{\frac{1}{N-1}\sum_{i=1}^{N}(NN_i - \mu_{NN})^2} \qquad (3)$$

The root mean square of the successive differences between NN intervals is,

$$\beta_{NN} = \sqrt{\frac{1}{N-1}\sum_{i=1}^{N}(NN_i - NN_{i+1})^2} \qquad (4)$$

Finally, the ability of the HR and HRV features to distinguish MCI and healthy classes are evaluated with the help of hypothesis testing and wrapper methods. In wrapper methods, the merit of the HR, and HRV features are assessed in terms of the accuracy of detecting the MCI observed on SVM, DA and NB classifiers. Wilcoxon rank-sum test is used to assesses the statistical significance of the features. Architecture of the proposed signal processing pipeline for computing HR & HRV features and validation discussed above is shown in figure 1.

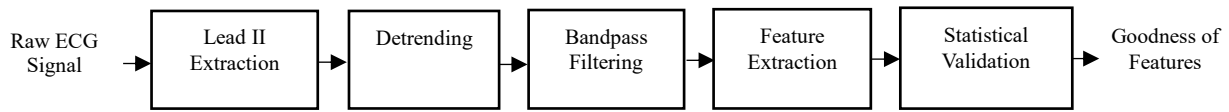

Figure 1: Architecture of the proposed signal processing pipeline for computing HR & HRV features and their validation

## 2.4 Statistical Validation of HR and HRV Features

We used Wilcoxon rank sum test to measure the extent to which the HR, and HRV features are different among MCI and non-MCI cases (Hwang et al., 2012). The Wilcoxon rank sum test is a non-parametric hypothesis test used to compare the features extracted from two independent groups when the data do not meet the assumptions of normality required by parametric tests like the t-test. It is commonly used when the data are ordinal, interval, or ratio scale but not normally distributed. The Wilcoxon rank sum test offers lower p-values significantly lesser than the level of significance when the distribution parameters of feature sets of different classes are unlike. In addition to the hypothesis test, we use wrapper methods features (Arco et al., 2024) also to assess the 'goodness' of HR, HRV features. The HR, and HRV features are assessed in terms of the accuracy offered by three ML models, namely SVM, DA and NB in distinguishing MCI, and non-MCI cases. A 10-fold cross validation strategy (Alharbi et al., 2022) is adopted in the ML-based validation. In the wrapper method, 70 % of the feature data is assigned as the training set and 30 % as the testing set.

## 3. Results

The p-values obtained from Wilcoxon rank-sum test on HR, and HRV features of MCI patients and healthy individuals are shown in Table 1. Bar graphs illustrating the p-values of the four different features Mean, and RMS, of NN intervals SDNN, RMSSD are shown in figure 2. All four features mean, and RMS, of NN intervals SDNN, RMSSD show p-values less than the level of significance 0.05. Low p-values convey that HR, and HRV features are different among MCI and non-MCI groups. Among the features, mean, RMS, of the NN intervals, are more significant than SDNN, and RMSSD implying that HR is more determinant than HRV features in distinguishing MCI, and non-MCI groups.

Table 1: p-values obtained from Wilcoxon rank-sum test on HR, and HRV features of MCI patients and healthy individuals

| Feature | Rank Sum Test (p-Value) |
|---|---|
| **Mean** | 0.0021 |
| **RMS** | 0.0014 |
| **SDNN** | 0.0192 |
| **RMSSD** | 0.0206 |

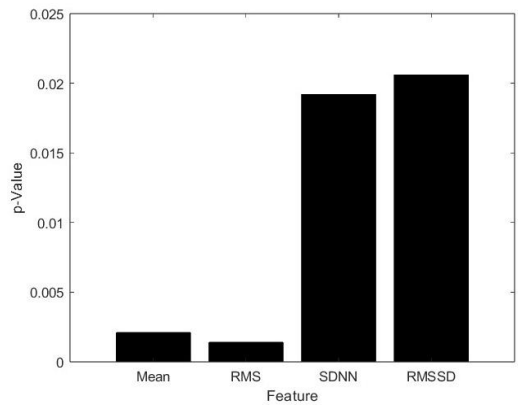

Figure 2: A bar graph illustrating the p-values of four different features: Mean of NN intervals, RMS of NN intervals, SDNN, and RMSSD

Figure 3 below shows the box whisker plots demonstrating the separability between the features, mean, and RMS of the NN intervals, SDNN and RMSSD. We can clearly see the nonoverlapping blocks of non-MCI and MCI groups for mean and RMS of NN intervals. Nonoverlapping blocks, and unequal median lines of the boxes of mean and RMS of the NN intervals of MCI and non-MCI classes justify that HR features are significantly different between the two groups.

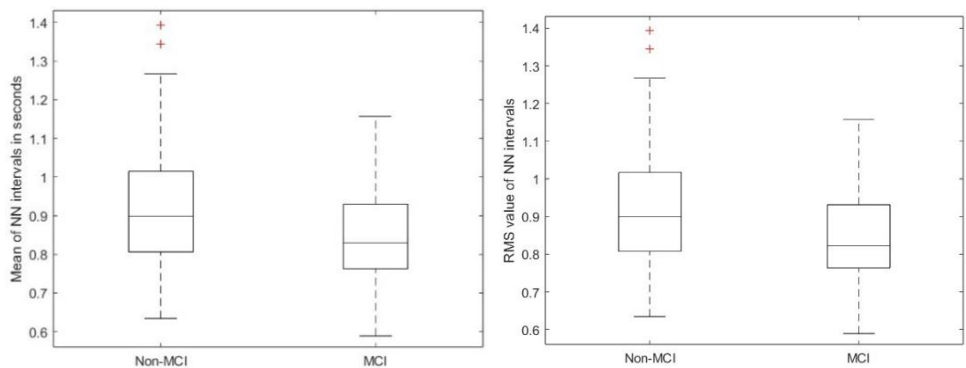

(a) (b)

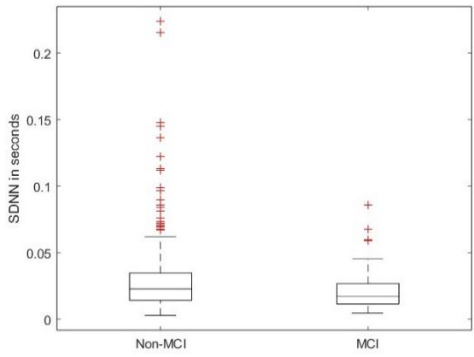
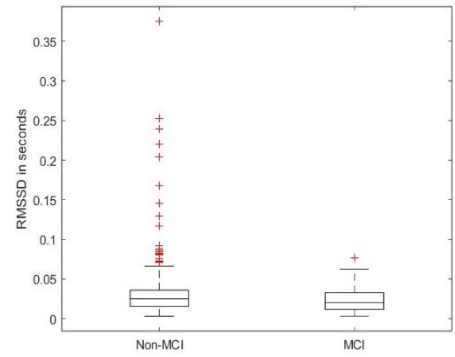

(c)                  (d)

Figure 3: Box Whisker plots showing the separability between features: (a) Mean of NN Intervals, (b) RMS of NN Intervals, (c) SD of NN intervals, and d) RMSSD of NN intervals

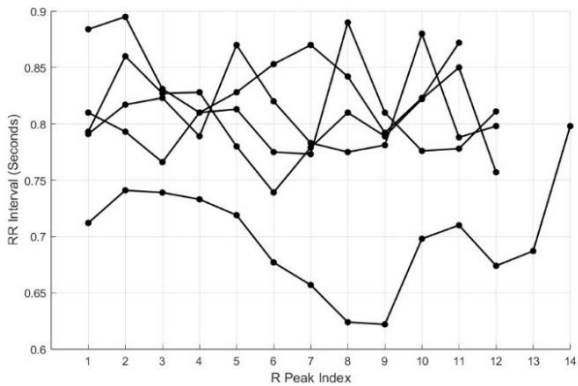
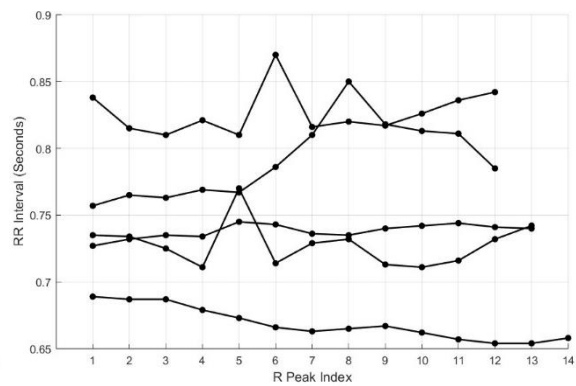

(a)                  (b)

Figure 4: RR intervals of (a) 5 non-MCI and (b) 5 MCI participants

Figure 4 shows the consecutive RR interval time series of 5 participants from MCI and non-MCI groups. For MCI patients, the RR interval is less, demonstrating that they have higher heart rate. Whereas, for non-MCI patients, the RR interval is more, indicating that the heart rate is comparatively less. It is also seen that the RR intervals of the non-MCI subjects exhibit greater variability compared to the RR interval values of the MCI patients. This indicates that the non-MCI subjects exhibit increased HRV.

Table 2 below shows the k-fold validation accuracies of mean, and RMS of NN intervals, SDNN and RMSSD for three different machine learning algorithms, namely, SVM, DA, and NB. Only one feature is used as input to a specific classifier at a time. All four features produce a very high accuracy of 80.80%, regardless of the classifier. The high accuracy values justify the potential of HR and HRV features to be used as a biomarker for detecting MCI.

Table 2: k-fold cross-validation accuracies for various machine learning classifiers.

| Feature | Support Vector Machine (SVM) Accuracy | Discriminant Analysis (DA) Accuracy | Naive Bayes (NB) Accuracy |
|---|---|---|---|
| Mean | 80.8081% | 80.8081% | 80.8081% |
| RMS | 80.8081% | 80.8081% | 80.8081% |
| SDNN | 80.8081% | 80.8081% | 80.8081% |
| RMSSD | 80.8081% | 80.8081% | 80.8081% |

**4. Discussion**

Our research highlights the potential of HR and HRV as a biomarker for MCI detection among the elderly Indian population. Notably, our study is one of a kind on its focus on examining HR and its variability for MCI detection within an Indian population. It is also the first study done in an elderly Indian population cohort, utilizing an ECG recording of only 10 seconds. Our findings align significantly with prior investigations that have explored the role of HR (Molloy et al., 2023), evaluated by the measures of central tendencies of NN intervals (mean and RMS), which can act as a biomarker for cognitive decline.

The literature demonstrates a connection between HR, HRV and cognition through the rank sum test and RE measure. These two methods displayed significant values as they showed a greater dissimilarity between the features, making it easier to distinguish the MCI group from the non-MCI group. This can be interpreted from the box whisker plots.

A comprehensive systematic review exploring the correlation between HRV and MCI found a prevailing positive relationship (Derby et al., 2023) across 19 out of 20 studies, signifying those individuals with higher HRV generally exhibited higher cognitive performance. This suggests a potential link between HRV and MCI. Notably, the variance in the outcomes of our study might be attributed to the duration of the ECG signal considered; we opted for shorter durations of ECG signal, of 10 seconds, while most of the other studies employed a minimum of 5-minute ECG signal duration (Forte et al., 2019). This temporal distinction could contribute to the observed disparities in our study outcomes compared to prior research.

In a study conducted by Schaich et. al. (2020), higher SDNN was associated with higher scores on the Cognitive Abilities Screening Instrument (CASI) and Digit Symbol Coding test (DSC) with p =

0.018 and p = 0.013 respectively. But there was no connection between HRV change and any of the cognitive scores. Here, we are showing a result which suggests that MCI has an influence on HR and HRV. The mean and RMS values in non-MCI are very high as seen from the box whisker plots. This signifies that the HR in such individuals without MCI will be low, whereas the HR in MCI patients will be high. Our observations are in accordance with the results of the study conducted by Imahori et al. (2022). This proves that HR is a major criterion that will distinguish individuals with MCI from those without MCI. Similarly, there is a greater HRV in patients with MCI than those without MCI. So, both HR and HRV can be used as biomarkers for the detection of MCI.

The major strengths of this study are: Here, we have used the ACE-III Cognitive Battery which is culturally validated in the Indian population to detect MCI. Demented patients were not included in this study; so, we demonstrated the ability of HR and HRV features to detect MCI in its pre-clinical stage itself. The use of short-term ECG recordings makes it convenient for clinical utility and point-of-care applications. In addition to the usual hypothesis tests, we used ML models also, to validate the HR and HRV features.

The current study did not include the transform-domain features like the power-ratio were not included since, we used ECG signals of short duration. The future work can focus on computing the transform-domain features without compromising the frequency resolution and signal quality. Also, combining the clinical parameters along with HR and HRV features could improve the diagnostic accuracy of MCI.

**5. Conclusion and Future Scope**

We investigated the effect of MCI on HR and HRV in the Indian population, using very short ECG recordings of individuals with MCI and those without MCI, for clinical utility and point-of-care applications. We designed a signal processing pipeline to detect the R-peaks from the pre-processed ECG signal and extract measures of central tendency (Mean and RMS) and dispersion (SDNN and RMSSD) of beat-to-beat intervals.

We obtained very small p-values for the mean, RMS, SDNN and RMSSD, which are very less than the level of significance, conveying the effect of MCI on HR and HRV. This makes it easier for the identification of features that distinguish MCI class from non-MCI class. People with MCI have increased HR and lower HRV when compared with healthy individuals. We assessed the discriminating capability of features with the help of various machine learning classifiers namely, SVM, DA and NB. The k-fold cross-validation accuracies were high for all three classifiers, proving the significance of HR and HRV for MCI detection. The suitability of the proposed method for the early detection of MCI and point-of-care applications along with the use of ACE-III battery are some of the major advantages of our study.

**Ethical Consideration**

The Institutional Ethics Committee approved the study protocol for the cohort. Written informed consent was taken and anonymity, privacy, and confidentiality were maintained for all the participants.

**Conflict of Interest**

Authors have no conflict of interest.

**Data Availability Statement**

The datasets generated and/or analysed during the current study are not publicly available as the study is a longitudinal cohort study and is currently ongoing, the data is still being collected and curated and being monitored by the Institutional Ethics Committee (IEC) and Technical Advisory Committee (TAC). Therefore, it is not made public at this point of time. Data request can be directed to the corresponding author Dr. Thomas Gregor Issac who is the PI of TLSA study and data will be shared if approved by the IEC and TAC.